\documentclass[runningheads]{llncs}
\usepackage{tikz}

\usepackage{hyperref}
\definecolor{lightgray}{gray}{0.9}
\usepackage{colortbl}
\usepackage{color}
\usepackage{float} 
\usepackage{graphicx}
\usepackage{booktabs} 
\usepackage{multirow} 
\usepackage{bbm} 
\usepackage{amssymb}
\usepackage{wrapfig} 
\usepackage{array}
\usepackage{graphicx}
\usepackage{subcaption}
\usepackage{marvosym}
\usepackage[marginal]{footmisc}

\usepackage{bm}
\usepackage{amsmath}
\usepackage{hyperref}
\usepackage{grffile}
\usepackage{pifont} 
\usepackage{mathtools} 
\usepackage{tabularx}
\usepackage{float}

\begin{document}
\setlength{\textfloatsep}{10pt}
\title{SurgTrack: CAD-Free 3D Tracking of Real-world Surgical Instruments}
\titlerunning{SurgTrack}
\author{Wenwu Guo\inst{1}{*}, Jinlin Wu\inst{1,2}{*}, Zhen Chen\inst{1}\textsuperscript{\Letter}, Qingxiang Zhao\inst{1}, Miao Xu\inst{1}\\Zhen Lei\inst{1,2,3}, Hongbin Liu\inst{1,2}}
\authorrunning{W. Guo et al.}
\institute{Centre for Artificial Intelligence and Robotics (CAIR), HKISI-CAS \and
MAIS, Institute of Automation, Chinese Academy of Sciences \and School of Artificial Intelligence, University of Chinese Academy of Sciences}

\maketitle             

\begin{abstract}
{
Vision-based surgical navigation has received increasing attention due to its non-invasive, cost-effective, and flexible advantages. 
In particular, a critical element of the vision-based navigation system is tracking surgical instruments. Compared with 2D instrument tracking methods, 3D instrument tracking has broader value in clinical practice, but is also more challenging due to weak texture, occlusion, and lack of Computer-Aided Design (CAD) models for 3D registration. To solve these challenges, we propose the SurgTrack, a two-stage 3D instrument tracking method for CAD-free and robust real-world applications. In the first registration stage, we incorporate an Instrument Signed Distance Field (SDF) modeling the 3D representation of instruments, achieving CAD-freed 3D registration. Due to this, we can obtain the location and orientation of instruments in the 3D space by matching the video stream with the registered SDF model. In the second tracking stage, we devise a posture graph optimization module, leveraging the historical tracking results of the posture memory pool to optimize the tracking results and improve the occlusion robustness. Furthermore, we collect the Instrument3D dataset to comprehensively evaluate the 3D tracking of surgical instruments. The extensive experiments validate the superiority and scalability of our SurgTrack, by outperforming the state-of-the-arts with a remarkable improvement. The code and dataset are available at \href{https://github.com/wenwucode/SurgTrack}{https://github.com/wenwucode/SurgTrack}}.
\end{abstract}

\section{Introduction}
Developing computer-assisted surgery systems can improve the quality of interventional healthcare for patients \cite{chen2023surgical,luo2023surgplan,chen2023temporal,bai2024endouic,xu2024transforming,chen2024asi,chen2024vs}, offering significant benefits, such as reduced operational times and minimized risk of surgical complications. In particular, surgical navigation systems have become an indispensable component in modern surgery  \cite{ma2023visualization,lehner2020site,van2018toward}, and ascertain the exact positioning of surgical instruments by tracking distinctive sections of the tools. As a critical element of surgical navigation systems, including electromagnetic-based \cite{shi2023research}, optical-based \cite{sorriento2019optical}, and vision-based systems \cite{yang2020review}. Among these, vision-based systems have garnered considerable interest due to non-invasive, cost-effective, flexible, and not subject to line-of-sight limitations or electromagnetic disturbances \cite{xu2023information,yang2020review}.

The 3D tracking algorithm is essential in vision-based surgical navigation systems \cite{sorriento2019optical}. However, most existing methods of instrument tracking are based on object-tracking algorithms, detecting the region of interest object and corresponding matching the detected region across different frames. Early works~\cite{journals/cars/ZhangYCY17} required markers of surgical instruments, and achieved instrument tracking by recognizing and matching markers across different frames. This method causes invasion of surgical instruments, lacking scalability. Later works \cite{journals/mia/BougetASJ17} proposed marker-freed tracking methods, detecting instruments with handcraft visual features and then tracking instruments through the Kalman filter algorithm. Limited by the generalizability of handcraft visual features, these marker-freed methods did not perform well in real-world applications. Recently, Fathollahi \textit{et al.} \cite{DBLP:conf/miccai/FathollahiSPDGA22} proposed a highly accurate instrument tracking method, which introduces Yolo-v5~\cite{glenn_jocher_2020_4154370} to improve the accuracy of instrument detection and applies ReID~\cite{ye2021deep} technology to improve the accuracy of cross-frame matching. However, these methods focused on developing 2D tracking of instruments, which can only perceive 2 degrees of freedom, which is not enough to provide sufficiently accurate information for surgical navigation.

Existing 2D tracking systems~\cite{journals/cars/ZhangYCY17} are restricted to the x and y planes, accommodating in-plane rotations for a total of three degrees of freedom. In comparison, 3D object tracking approaches~\cite{zhu2022nice,slavcheva2018sdf,newcombe2011kinectfusion,teed2021droid,wen2021bundletrack} match detected objects with pre-established computer-aided design (CAD) models to ascertain their 3D orientation. Represented through six degrees of freedom—spanning the x, y, and z axes, and including the rotational dimensions of pitch, yaw, and roll—this detailed spatial understanding is vital for vision-based navigational systems. However, the application of these 3D tracking methods to surgical environments is fraught with challenges. A primary challenge is the inaccessibility of CAD models for surgical instruments, as they are often proprietary due to patent protections. The absence of CAD models hinders most 3D tracking techniques in the realm of surgical instrument tracking. Additional obstacles are the low textural features and frequent occlusions of surgical instruments, which complicate their detection and sustained tracking.

Inspired by existing works~\cite{mildenhall2020nerf,slavcheva2018sdf,wen2023bundlesdf}, we design a novel 3D surgical instrument tracking method, named SurgTrack, which is capable of accurately tracking the 6 degrees of freedom of surgical instruments in real 3D space. To solve the problem of missing CAD models, we incorporate an Instrument Signed Distance Field (SDF) model generating the 3D representation of the surgical instrument with RGB-D video frames. We also propose an Instrument SDF model to further accurately learn the 3D shape and texture of instruments. Through Instrument SDF, SurgTrack completes the registration of 3D tracking without CAD models. To solve tracking problems caused by occlusion and weak textures, we apply a posture memory pool to provide historical tracking results as a reliable reference. We also utilize a posture graph optimization module to optimize the ongoing tracking results with historical references and ensure that occlusions and weak textures do not cause tracking interruptions.

Furthermore, to facilitate a comprehensive analysis and evaluation of our methods, we collect a 3D tracking dataset of surgical instruments, named Instrument3D. Our SurgTrack achieves a remarkable 3D tracking performance with the 88.82\% ADD-S and the 12.85 reconstruction error. We also conduct experiments on the general 3D object tracking dataset HO3D to demonstrate the generalization and scalability of our SurgTrack.

\section{Method}
\subsection{Overview of SurgTrack}\label{sec_overview}

\begin{figure}[t]
    \centering
    \includegraphics[width=1\linewidth]{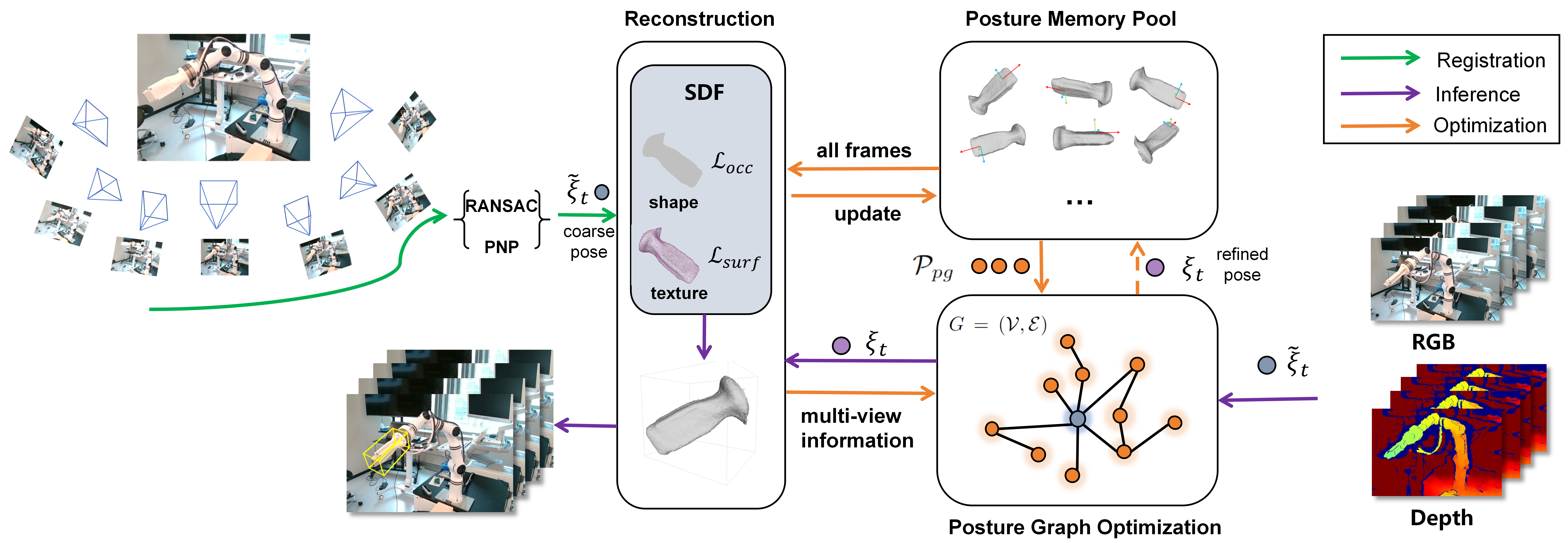}
    \caption{The overview of our SurgTrack framework. Our SurgTrack comprises a registration stage and a tracking stage. (a) In the registration stage, the 3D model is reconstructed using the instrument SDF. (b) The tracking stage first uses RANSAC for rough pose estimation, and then performs pose optimization using posture memory pool and posture graph.}
    \label{fig:enter-label}
\end{figure}

An overview of our SurgTrack framework is shown in Fig.~\ref{fig:enter-label}. To achieve CAD-free registration, we first model the 3D shape of the surgical instrument using SDF (\ref{sec_opt1}). Then, we track the 6-DoF pose of the instruments through the Posture Memory Pool and Posture Graph Optimization (\ref{sec_opt2}).

\subsection{CAD-free Instrument Registration}\label{sec_opt1}

\noindent\textbf{Instrument SDF Modeling.} 
Given the 3D point cloud $\{v| v \in \mathbb{R}^{3}\}$ captured by a RGB-D camera, we adapt the Signed Distance Function (SDF) to model the 3D representation of the surgical instrument as follows:
\begin{equation}
    S = \{ v | \Psi(v)=0 \},
    \label{sdf}
\end{equation}
where $\Psi(v)=0$ represents the points on the surface of the instrument. Therefore, we can derive the 3D model of the instrument from point cloud data, eliminating the need for a pre-existing Computer-Aided Design (CAD) model. This 3D model facilitates the registration process for 3D tracking. However, the SDF methodology faces inherent limitations when dealing with complex scenarios, such as occlusions and low-texture regions.

\noindent\textbf{Occlusion and Texture Optimization.} To address this, we incorporate the occlusion constraint and shape constraint in the SDF model. For occlusions, we introduce a positive value \(\delta\) to alleviate boundary ambiguities between background and instrument caused by partial occlusions:
\begin{equation}
   \mathcal{L}_{occ} = \frac{1}{|V_{occ}|}\sum_{v\in V_{occ}}(\Psi(v)-\delta)^{2}.\label{con:1}
\end{equation}
For surfaces with weak textures, we consider points near the surface in the SDF modeling process, enabling our SurgTrack to better capture the surface geometry and handle areas with weak textures, as follows:
\begin{equation}
   \mathcal{L}_{surf} = \frac{1}{|V_{surf}|}\sum_{v\in V_{surf}}\left(\Psi(v)+d_{v}-d_{\Delta}\right)^{2}.\label{con:2}
\end{equation}
In this way, the total loss function $\mathcal{L}$ is defined as follows:
\begin{equation}
    \mathcal{L} = \alpha \mathcal{L}_{occ} + \beta \mathcal{L}_{surf},\label{con:3}
\end{equation}
where $\alpha$ and $\beta$ balance the contributions of the two components.

\subsection{Instrument Tracking}\label{sec_opt2}

\subsubsection{Tracking Initialization.}

In the tracking stage, we estimate a coarse pose \(\tilde{\xi}_t\) by matching the current frame and its adjacent frames with RANSAC algorithm as follows:
\begin{equation}
\tilde{\xi}_t = \arg \min_{R, t} \sum_{i} \left\| R p_i + t - q_i \right\|^2.
\end{equation}
In the above equation, RANSAC algorithm minimizes the distance between the reconstructed results \(p_i\) and their corresponding scene points \(q_i\) and estimates the coarse pose $\tilde{\xi}_t$. $R$ and $t$ represent the rotation and translation matrix.

\subsubsection{Tracking Optimization.}

Following the initial rough pose estimation obtained using RANSAC, the pose \(\tilde{\xi}_t\) serves as the initial estimate in the subsequent optimization phase. This pose is further refined by integrating the pose memory pool with the pose graph to improve accuracy and robustness. First, to address challenges such as long-term tracking drift, data loss, and occlusions, it is crucial to preserve the pose data from previous frames. We implement a posture memory pool \(\mathcal{P}\) that stores this information as follows:
\begin{equation}
\mathcal{P} = \{ (\xi_i, M_i) \mid i = 1, 2, \ldots, N \},    
\end{equation}
where \(\xi_i \in \text{SE}(3)\) represents the optimized pose of the \(i^{th}\) frame, \(M_i\) contains the 3D point cloud data associated with the \(i^{th}\) frame, and \(N\) is the number of keyframes currently stored in the posture memory pool.

With the initial pose \(\tilde{\xi}_t\), we construct a posture graph using selected relevant frames from the posture memory pool. The selection is based on criteria such as the RANSAC matching threshold and frame overlap to ensure reliable references. Then, the posture graph is constructed as follows:
\begin{equation}
    G=(\mathcal{V},\mathcal{E}),
\end{equation}
where the nodes $\mathcal{V}$ consist of the current frame \(\mathcal{F}_t\) and the selected reference frames $\mathcal{P}_{pg}$, as \(\mathcal{V}=\mathcal{F}_t\cup\mathcal{P}_{pg}\) with \(|\mathcal{V}|=K+1\).

Based on the posture graph, we refine the tracking results of the current frame through the following loss function, resulting in the final optimized pose \(\xi_t \in \text{SE}(3)\):
\begin{equation}
\xi_t \leftarrow \arg \min_{\xi_t} \left(w_s\mathcal{L}_{\text{SDF}}(t)+\sum_{i\in\mathcal{V},j\in\mathcal{V},i\neq j}\left[w_f\mathcal{L}_{\text{3D}}(i,j)+w_p\mathcal{L}_{\text{2D}}(i,j)\right]\right),    
\end{equation}
where \(\mathcal{L}_{\text{3D}}(i,j)\) is the 3D distance loss, \(\mathcal{L}_{\text{2D}}(i,j)\) is the 2D projection loss, \(\mathcal{L}_{\text{SDF}}(t)\) is the instrument SDF depth loss, and the scalar weights \(w_f,w_p,w_s\) are empirically set to 1. Specifically, the 3D distance loss $\mathcal{L}_{\text{3D}}$ is calculated as:
\begin{equation}
\mathcal{L}_{\text{3D}}(i,j)=\sum_{(p_m,p_n)\in C_{i,j}}\rho\left(\left\|\xi_i^{-1}p_m-\xi_j^{-1}p_n\right\|_2\right).    
\end{equation}
This 3D distance loss measures the Euclidean distance between corresponding RGB-D features \(p_m,p_n\in\mathbb{R}^3\), using the Huber loss function \(\rho\) to enhance the robustness of our SurgTrack.

On the other hand, the 2D projection loss $\mathcal{L}_{\text{2D}}$ is calculated as:
\begin{equation}
\mathcal{L}_{\text{2D}}(i,j)=\sum_{p\in I_i}\rho\left(\left|n_i(p)\cdot\left(T_{ij}^{-1}\pi_{D_j}^{-1}(\pi_j(T_{ij}p))-p\right)\right|\right).    
\end{equation}
This 2D projection loss assesses the pixel-wise point-to-plane distance after projection and transformation, comparing node \(i\) to the plane in node \(j\).

Finally, the instrument SDF depth loss $\mathcal{L}_{\text{SDF}}$ is calculated as follows:
\begin{equation}
\mathcal{L}_{\text{SDF}}(t)=\sum_{p\in I_{t}}\rho\big(\left|\Psi(\xi_t^{-1}(\pi_{D}^{-1}(p)))\right|).   
\end{equation}
This instrument SDF depth loss measures the distance between the current frame and the implicit surface defined by the Instrument SDF, where \(\Psi(\cdot)\) is the signed distance function indicating proximity to the surface. Note that this loss is considered only after the initial training of the object field has converged.

In this way, the optimization strategy for our SurgTrack, starting from the rough pose \(\tilde{\xi}_t\) and resulting in the final optimized pose \(\xi_t \in \text{SE}(3)\), integrates 3D spatial information, instrument shape, and depth data from a single viewpoint to complete pose optimization, improving robustness against reflections, weak textures, and long-term tracking challenges.

\section{Experiments}

\subsection{Experimental Settings}
\textbf{Datasets.} We collect a 3D tracking dataset of surgical instruments in RGB-D videos, named Instruments3D. The Instruments3D dataset consists of 13 videos across 5 surgical instruments, including ultrasound bronchoscopes, flexible and rigid endoscopes, thoracoscopes, and ultrasound probes. The Instruments3D dataset presents RGB-D videos by capturing human hands manipulating YCB objects, recorded at close range using an Intel RealSense camera. The ground truth data is derived through multi-view registration. We also conduct experiments on the general object 3D tracking dataset, HO3D \cite{hampali2020honnotate,hampali2022keypointtransformer}.

\noindent\textbf{Evaluation metrics.} We follow the classical evaluation protocol of 3D object tracking \cite{hampali2020honnotate,hampali2022keypointtransformer}. We use the ADD and ADD-S as the accuracy metric of 3D tracking, with their values ranging from 0 to 1, where higher values signify better accuracy. We use the Chamfer Distance (CD) as a measure of reconstruction error, where a smaller value indicates a more precise reconstruction.

\begin{figure}[t]
    \centering
    \includegraphics[width=1\linewidth]{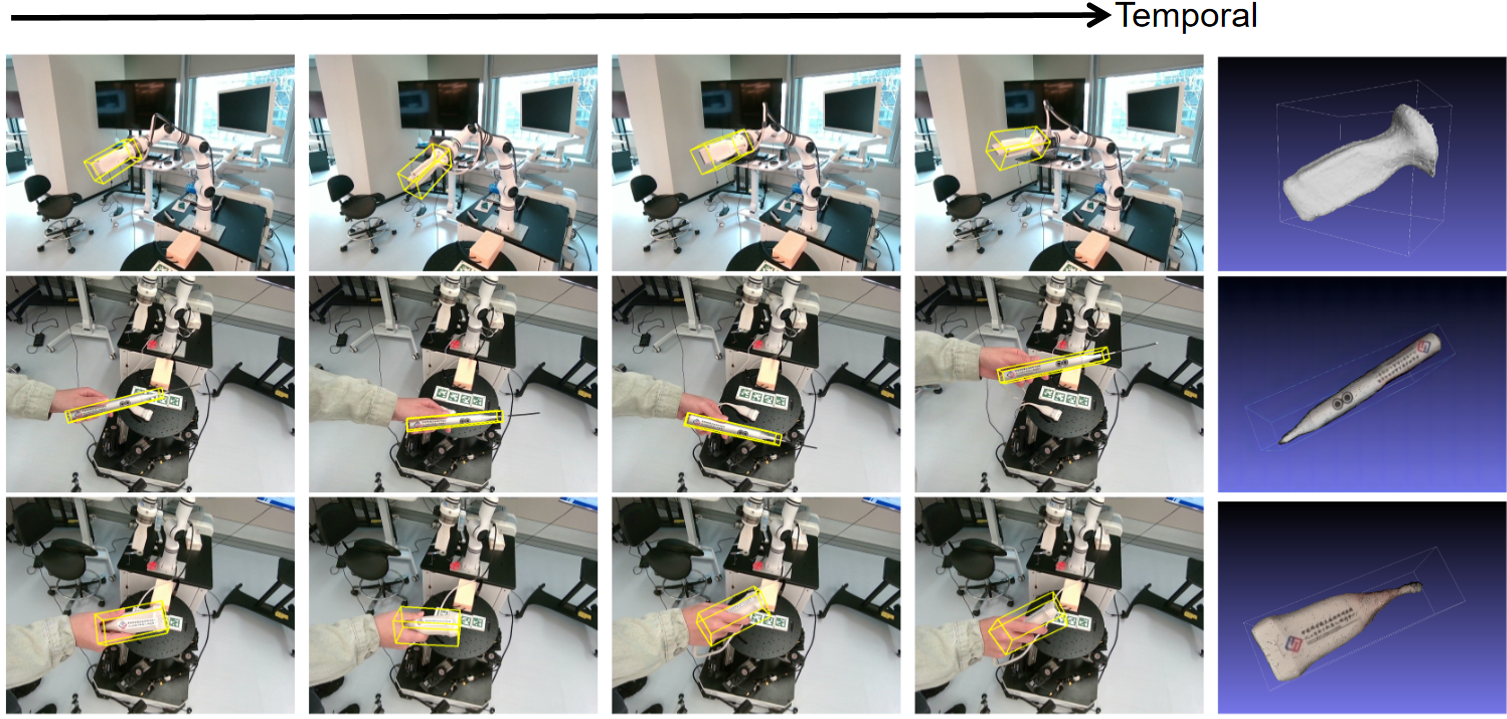}
    \caption{Visualization on the Instrument3D dataset test samples. Our SurgTrack achieves causal 3D tracking and reconstruction of weakly textured surgical instruments for monocular RGB-D sequences.}
    \label{fig:four-images-instrument}
\end{figure}

\subsection{Comparison Results on Instrument3D and HO3D}\label{sec_opt}

\begin{table}[t]
    \centering
    \begin{tabular}{c|c|c c|c}
    \toprule
    \multirow{2}{*}{Dataset} & \multirow{2}{*}{Method} & \multicolumn{2}{c|}{Pose} & \multicolumn{1}{c}{Reconstruction} \\ 
    & & $\text{ADD-S }(\%)\uparrow$ & $\mathrm{ADD}\left(\%\right)\uparrow $ & $\mathrm{CD}\left(\mathrm{cm}\right)\downarrow $\\
    \midrule \midrule
    \multirow{5}{*}{HO3D} & NICE-SLAM \cite{zhu2022nice} & 22.29 & 8.97 &52.57\\
    & SDF-2-SDF \cite{slavcheva2018sdf} & 35.88 & 16.08 &9.65\\
    & KinectFusion \cite{newcombe2011kinectfusion} & 25.81 & 16.54 &15.49\\
    & DROID-SLAM \cite{teed2021droid} & 64.64 & 33.36 & 30.84\\
    & BundleTrack \cite{wen2021bundletrack} & 92.39 & 66.01 & 52.05\\
    \rowcolor{lightgray} & SurgTrack & 95.85 & 92.53 &  0.65\\
    \bottomrule
    \midrule
    \multirow{3}{*}{Instrument3D} & Pixtrack \cite{chidananda2022pixtrack} & 60.59 & 45.13 &- \\
    & OnePose \cite{sun2022onepose} & 30.06 & 16.98 & - \\
    \rowcolor{lightgray} & SurgTrack & 88.82 & 83.65 &  12.85\\
    \bottomrule
    \end{tabular}
    \vspace{0.5em} 
    \caption{Comparison of our SurgTrack with state-of-the-arts on the Instrument3D and HO3D datasets. The ADD and ADD-S are AUC percentages (0 to 0.1 m). Reconstruction is measured by Chamfer Distance (CD).}
    \label{tab:merged_comparative_results}
\end{table}

\noindent\textbf{Comparison on Instrument3D.}
The Instrument3D dataset presents a complex challenge due to the frequent occlusions and severe motion blur encountered during the manipulation of surgical instruments. Furthermore, the inherent characteristics of these instruments such as their weak texture, reflective surfaces, and slender profiles compound the difficulty. Despite these difficulties of the Instrument3D dataset, our SurgTrack maintains the capability of robust, long-term tracking in most cases, as shown in Fig.~\ref{fig:four-images-instrument}. The comparison results in Table~\ref{tab:merged_comparative_results} confirm the remarkable advantage of our SurgTrack over state-of-the-art 3D tracking methods.

\begin{figure}[t]
    \centering
    \includegraphics[width=1\linewidth]{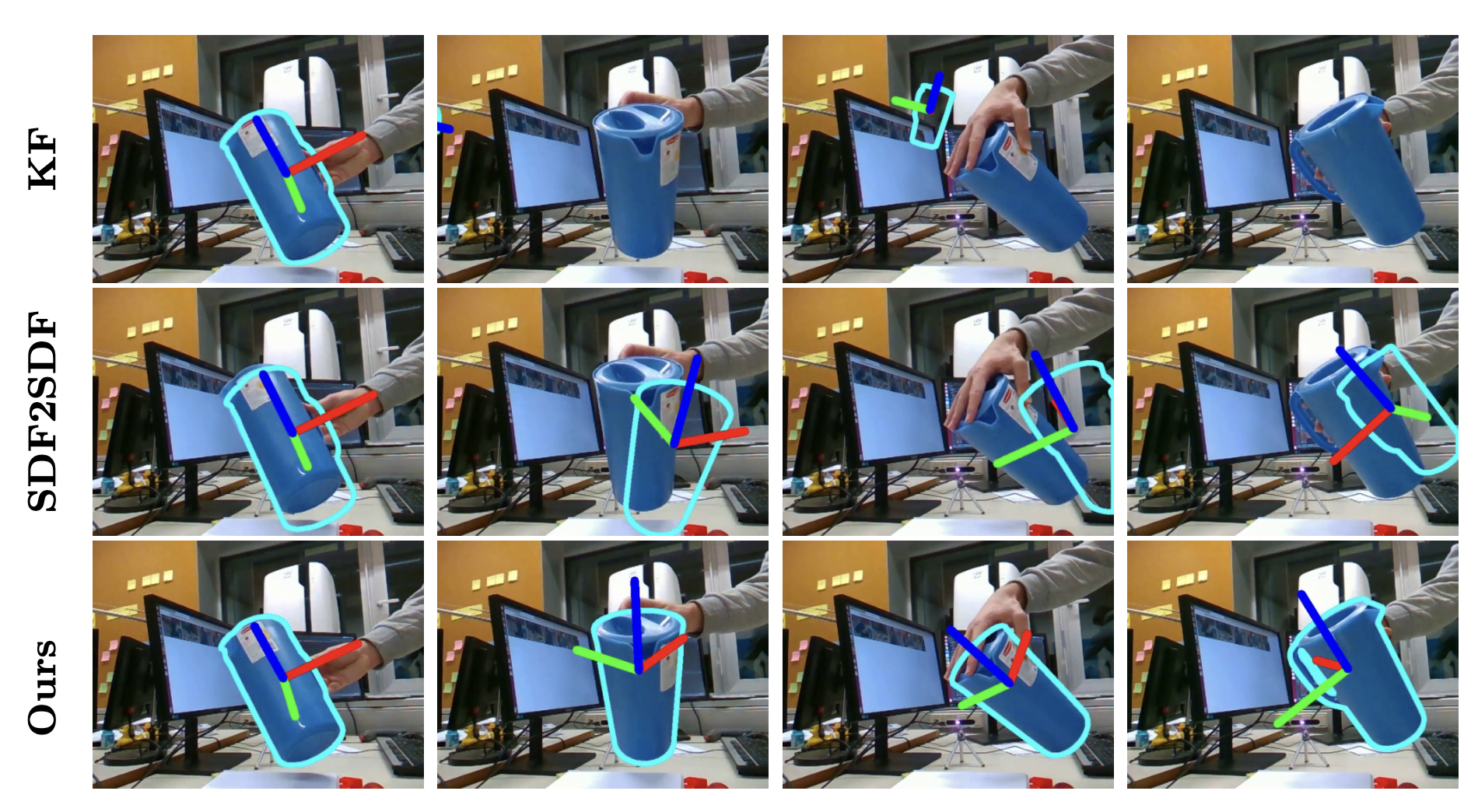}
    \caption{The 3D tracking visualization of our SurgTrack and state-of-the-art methods (\textit{i.e.}, KF and SDF-2-SDF) on the HO3D dataset.}
    \label{fig:hodpic}
\end{figure}

\noindent\textbf{Comparison on HO3D.}
As shown in Table~\ref{tab:merged_comparative_results} and Fig.~\ref{fig:hodpic}, on the HO3D dataset, we achieve the best results compared with other tracking schemes. Our algorithm shows strong capabilities in both ADD-S and ADD. While BundleTrack matches our performance in ADD-S, it falls short in all other metrics where we excel significantly, and it also demands more than 300 rounds of training. This demonstrates the strong generalization ability of our method to general objects.

\subsection{Ablation Study}

\begin{table}[t] 
    \centering
    \begin{tabular}{l|c c|c}
    \toprule
    \multirow{2}{*}{Ablations} & \multicolumn{2}{c|}{Pose} & \multicolumn{1}{c}{Reconstruction} \\ 
    & $\text{ADD-S }(\%)\uparrow$ & $\mathrm{ADD}\left(\%\right)\uparrow $ & $\mathrm{CD}\left(\mathrm{cm}\right)\downarrow $\\
    \hline \hline
    \rowcolor{lightgray} SurgTrack & 88.82 & 83.65 &  12.85\\
    \quad \textit{w/o} occlusion and texture optimization & 76.39 &  62.09 & 35.52\\
    \quad \textit{w/o} posture memory pool & 75.65 & 47.14 &-\\
    \quad \textit{w/o} posture graph & 77.23 &  42.36 & 15.60\\
    \bottomrule
    \end{tabular}
    \vspace{0.5em} 
    \caption{Ablation study of our SurgTrack on the Instrument3D dataset.}
    \label{tab:comparative_results}
\end{table}

To comprehensively evaluate our SurgTrack framework for 3D tracking of surgical instruments, we investigate the impact of each module. These modules include occlusion and texture Optimization, posture memory pool, and posture graph. As shown in Table~\ref{tab:comparative_results}, the occlusion and texture optimization is helpful for tracking optimization, which can increase ADD-S by 12.43\% and ADD by 21.56\%. When constructing the posture graph, selecting the most matching pose subset instead of randomly selecting can reduce the CD error by nearly 3cm and increase the ADD by 41.29\%. In this way, these comparisons further validate the effectiveness of our SurgTrack with tailored modules.

\section{Conclusion}
In this study, we collect a new multi-category surgical instrument 3D tracking data set, conduct a comprehensive study on 3D surgical instrument tracking, and propose a framework for 3D instrument tracking. We use Instrument SDF to generate the 3D representation of surgical instruments, achieving CAD-free 3D tracking registration. In the tracking stage, we use the posture memory pool and combine it with the posture graph for pose optimization, which greatly improves the 3D tracking accuracy. We also use the Instrument SDF to further improve the robustness to occlusion, weak texture, and long-term tracking. Experiments show that our method has significant superiority and scalability over public data sets and surgical instrument 3D tracking datasets.

\begin{credits}
\subsubsection{\ackname} This work was supported by the  National Natural Science Foundation of China (Grant No.\#62306313 and No.\#62206280), and the InnoHK program.

\subsubsection{\discintname}
The authors declare no competing interests.
\end{credits}

%
%
%
\bibliographystyle{splncs04}
\bibliography{mybibliography}
\end{document}